\documentclass[10pt,twocolumn,letterpaper]{article}

\usepackage{cvpr}
\usepackage{times}
\usepackage{epsfig}
\usepackage{graphicx}
\usepackage{amsmath}
\usepackage{amssymb}
\usepackage{enumitem}
\usepackage{subcaption}
\usepackage{float}
\usepackage{mathtools, nccmath}
\usepackage[export]{adjustbox}

\usepackage[pagebackref=true,breaklinks=true,letterpaper=true,colorlinks,bookmarks=false]{hyperref}

\cvprfinalcopy 

\def\cvprPaperID{39} 

\ifcvprfinal\fi
\begin{document}

\title{Uncertainty Gated Network for Land Cover Segmentation}

\author{Guillem Pascual\\
University of Barcelona\\
{\tt\small guillem.pascual@ub.edu}
\and
Santi Segu\'i\\
University of Barcelona\\
{\tt\small santi.segui@ub.edu}
\and
Jordi Vitri\`a\\
University of Barcelona\\
{\tt\small jordi.vitria@ub.edu}
}

\maketitle

\begin{abstract}
   The production of thematic maps depicting land cover is one of the most common applications of remote sensing. To this end, several semantic segmentation approaches, based on deep learning, have been proposed in the literature, but land cover segmentation is still  considered an open problem due to some specific problems related to remote sensing imaging. In this paper we propose a novel approach to deal with the problem of modelling multiscale contexts surrounding pixels of different land cover categories. The approach leverages the computation of a heteroscedastic measure of uncertainty when classifying individual pixels in an image. This classification uncertainty measure is used to define a set of memory gates between layers that allow a principled method to select the optimal decision for each pixel. 
\end{abstract}

\section{Introduction}

Land cover segmentation deals with the problem of multi-class semantic segmentation of remote sensing images. This problem, which consists of assigning a unique label (or class) to every pixel of an image, is particularly difficult due to (i) the high resolution of the images and diversity of size of the objects, (ii) the diversity of classes and, usually, the similarities among them, (iii) the noisy labeling and implicit rules such as not considering small/isolated areas and (iv) data domain: the model is usually trained with a set of images that highly differ from the target area where it is expected to generalize and perform predictions. 

\begin{figure}[h!]
    \captionsetup[subfigure]{labelformat=empty}
    \centering
    \begin{adjustbox}{minipage=\columnwidth,scale=0.7}
    \begin{subfigure}[t]{0.49\columnwidth}
        \raisebox{-\height}{\includegraphics[width=\textwidth]{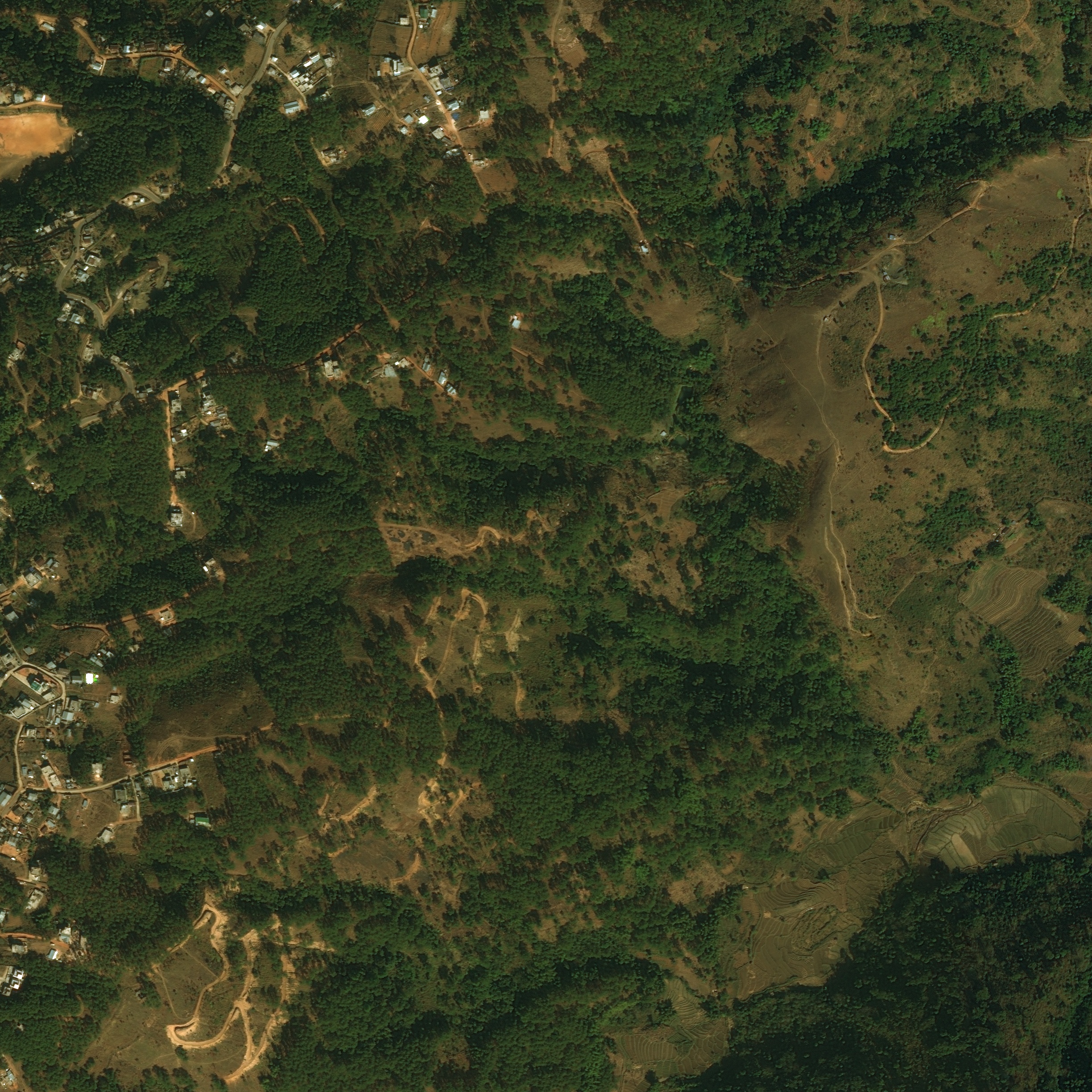}}
        \caption{~}
    \end{subfigure}
    \hfill
    \begin{subfigure}[t]{0.49\columnwidth}
        \raisebox{-\height}{\includegraphics[width=\textwidth]{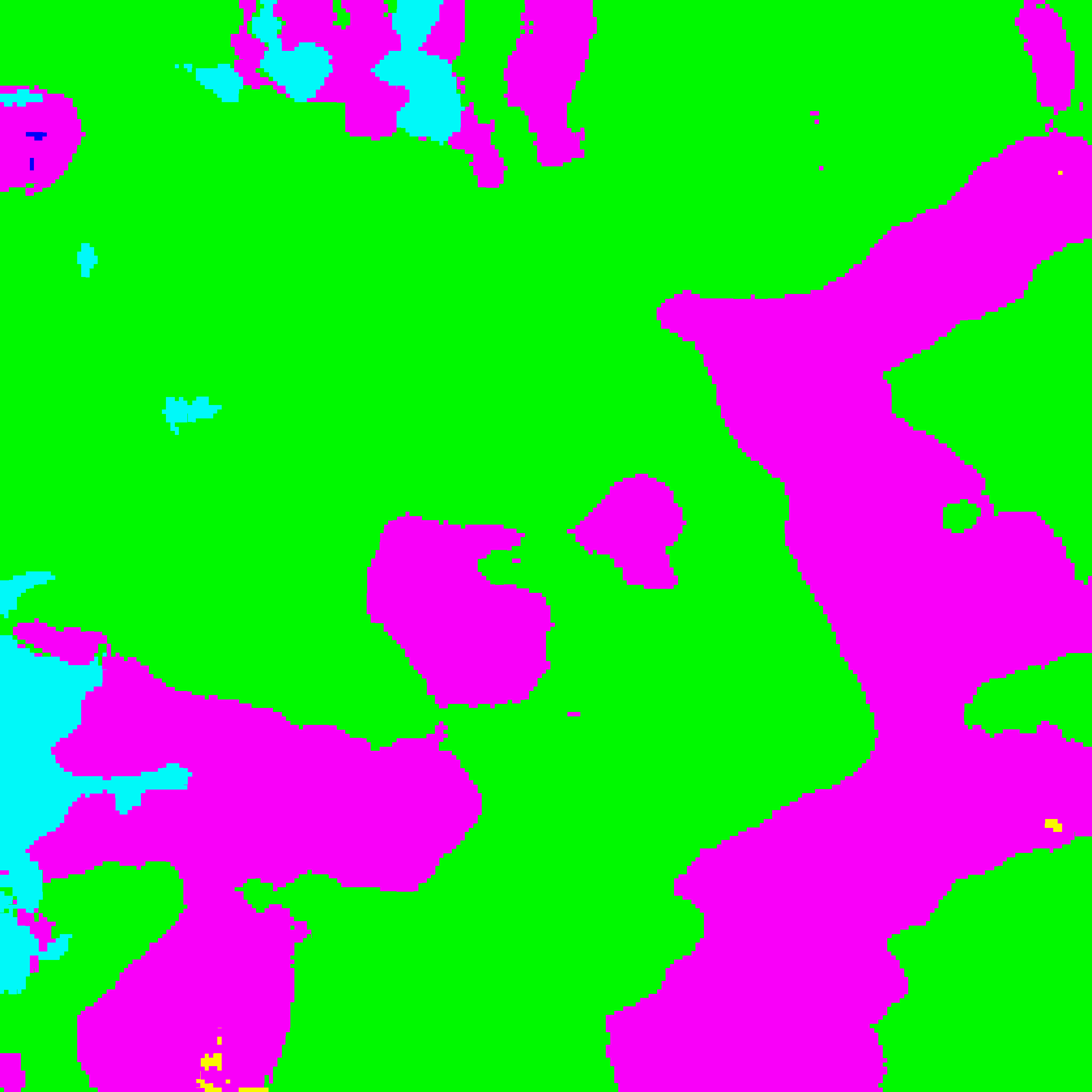}}
        \caption{~}
    \end{subfigure}
    \begin{subfigure}[t]{0.49\columnwidth}
        \raisebox{-\height}{\includegraphics[width=\textwidth]{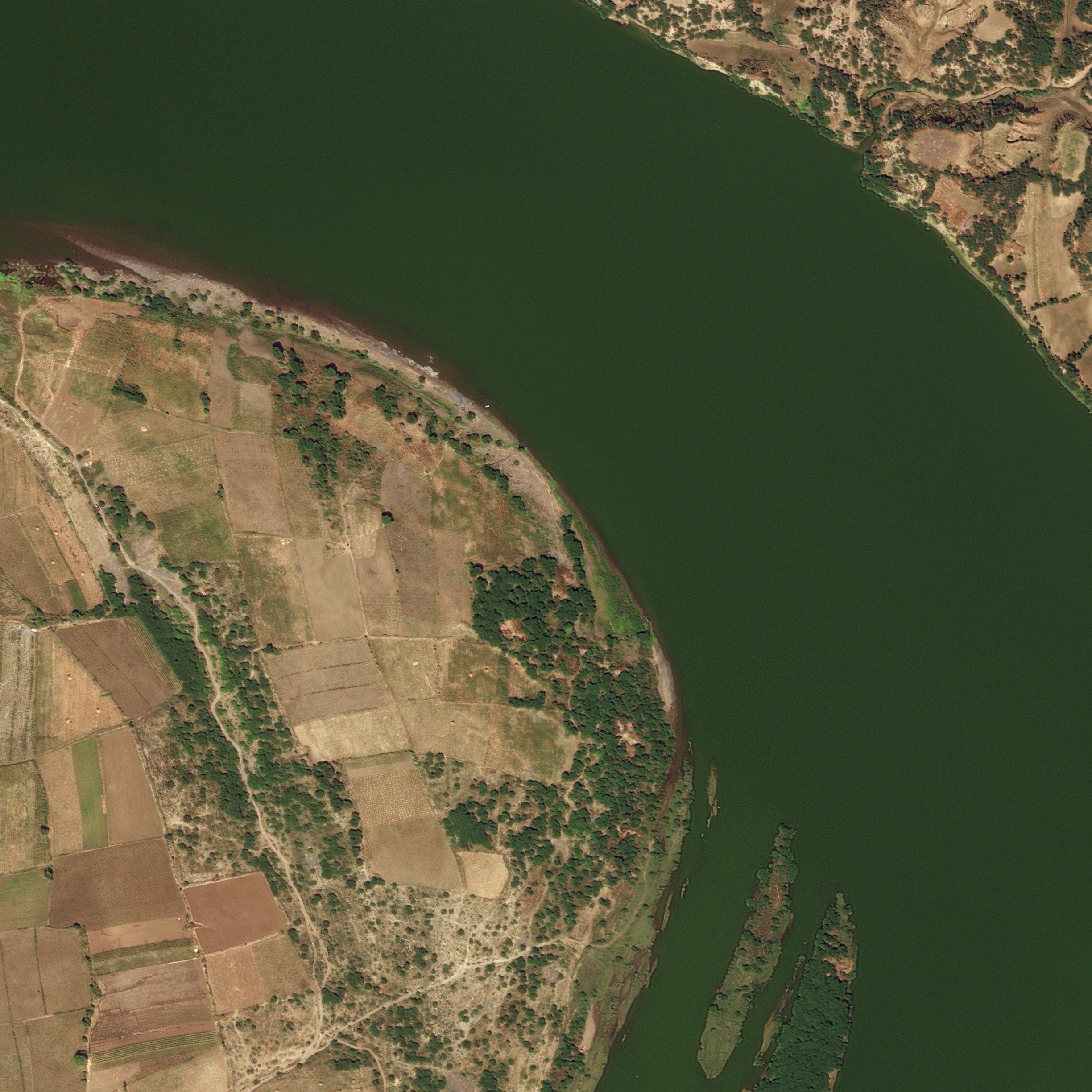}}
    \end{subfigure}
    \hfill
    \begin{subfigure}[t]{0.49\columnwidth}
        \raisebox{-\height}{\includegraphics[width=\textwidth]{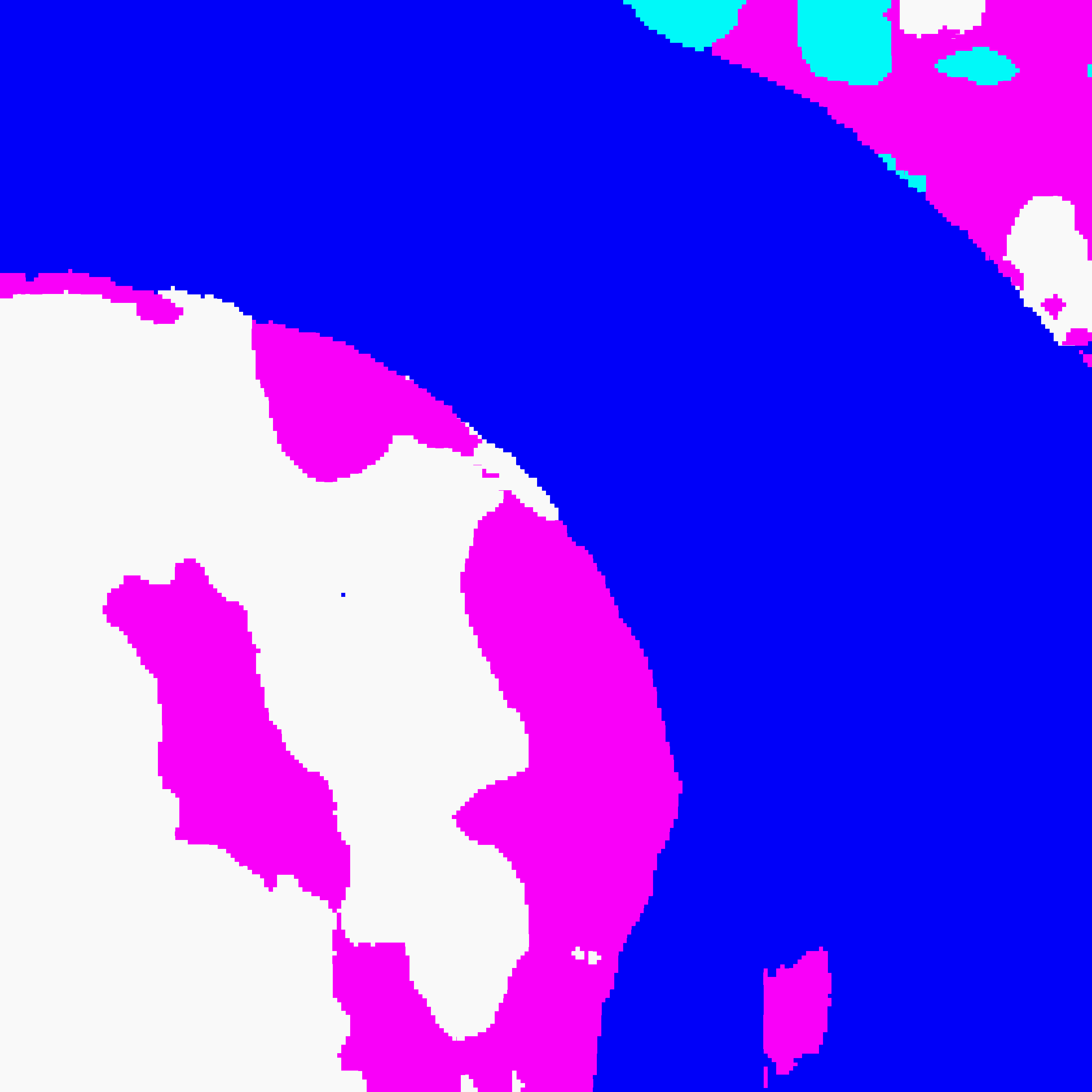}}
    \end{subfigure}
    \end{adjustbox}
    \caption{Prediction examples from the model. Each row is a sample, the left column is the input image and the right column is the predicted segmentation map.}
    \label{fig:example}
\end{figure}

In the recent literature, most of the methods solving these problems are based on deep learning. In \cite{long2015fully} Long \etal popularized the use of fully convolutional networks for segmentation. This method without any dense layer, allowed to create segmentation maps for images of any size. Based on this idea, and also trying to solve the exact alignment problem associated with the pooling layers, several methods have been presented \cite{8014890,unet,zhao2017pspnet,lin2017refinenet}. U-Net \cite{unet} is a popular architecture defined as an encoder-decoder scheme, where in the encoder stage the spatial dimension is gradually reduced with pooling layers and then decoder stage gradually recovers the object details and spatial dimension to finally obtain the output segmentation map. In  RefineNet \cite{lin2017refinenet}, proposed by Lin \etal, the ResNet architecture is used as a encoder step while in the decoder step as a set of RefineNet blocks which fuse high resolution features from the encoder and low resolution features from previous RefineNet block.

In the domain of satellite images, several methods trying to solve this problem in high-resolution images have been presented \cite{Sherrah16,rs9050446,2018arXiv180404020N,rs9060522}. The most relevant publication for our work is Gated Convolutional Network (GCN) \cite{rs9050446}, where the segmentation is computed from the outputs of each block of a pre-trained ResNet, using entropy as a gate to fine-tune the prediction at each level. 

In this paper, we propose a novel method that tackles the problem of land cover segmentation using the data and protocol proposed by the DeepGlobe Land Cover Classification Challenge at CVPRW \cite{demir2018deepglobe}. Figure \ref{fig:example} shows two samples from the dataset and the predictions of our model. The proposed method is built over a GCN using a ResNet architecture and exploits the uncertainty of the predictions in each layer. The uncertainty measure, built on the basis of the publication by Alex Kendall and Yarin Gal \cite{kendall2017uncertainties},  is used to define a set of memory gates between layers that allow for a principled method to select the optimal decision for each pixel.

The remainder of this paper is organized as follows. In the next section we present the proposed method. In Section III, we present the training setup. In Section IV, we present the experimental results. Finally, Section IV concludes the paper with remarks on the proposed approach.


\section{Method}

Our model builds upon the GCN architecture proposed by Wang \etal in \cite{rs9050446}. In that paper a new architecture was proposed to combine the feature maps learned at different blocks of a ResNet model by using memory gates instead of more classical operations such as summation or concatenation. The gating mechanism was based on the relationship between the information entropy of the feature maps and the label-error map, allowing for a better feature map integration.  

 To further develop the concept of gated convolutions, we consider the use of a more principled concept: assigning a credibility measure to each feature map. According to the Bayesian viewpoint proposed by Alex Kendall and Yarin Gal in \cite{kendall2017uncertainties}, it is possible to characterize the concept of uncertainty into two categories. On the one hand, if the noise applies to the model parameters, we will refer to \textit{epistemic uncertainty}. On the other hand, if the noise occurs directly in the observation, we will refer to it as \textit{aleatoric uncertainty}. Additionally, aleatoric uncertainty can further be categorized into two more categories: \textit{homoscedastic uncertainty}, when the noise is constant for all the outputs (thus acting as a ``measurement error"), or \textit{heteroscedastic uncertainty} when the noise of the output also depends explicitly on the specific input.
 
 We propose to use a measure of heteroscedastic uncertainty when classifying specific pixels as a gating mechanism. In this case, we have to measure the heteroscedastic uncertainty in a classification task, where the noise model is placed in the {\em logit} space. 
 
 Let $\sigma_i$ and $l_i$ be two predicted vectors of unaries of dimension $C$, the number of classes, for every input pixel $x_i$. The latter, $l_i$, are the logits used to output a probability distribution by using a softmax, while $\sigma_i$ aims to bound its uncertainty. By taking $T$ random samples of $l_i$ perturbed by $\sigma_i$, we can derive a stochastic loss $\mathcal L_x$ that allows the computation of an uncertainty value $\gamma_i$ for each $x_i$ input as follows:
 
 $$
 \mathcal L_x = \sum_i \gamma_i
 $$
 $$
  \gamma_i = -\log \frac{1}{T} \sum_t \exp(\hat{l}_{i,t,c} - \log \sum_{c'} \exp \hat{l}_{i,t,c'})
 $$
 $$
  \hat{l}_{i,t} \sim \mathcal{N}(l_i, \sigma_i), 1\leq t \leq T
 $$
 
 where $\hat{l}_{i,t,c'}$ is the $t$ sampled logit vector from class $c'$,  and $\hat{l}_{i,t,c}$ is the logit vector of the winner class for each pixel and sample.
 
\begin{figure}
    \begin{center}
        \includegraphics[width=1.0\columnwidth]{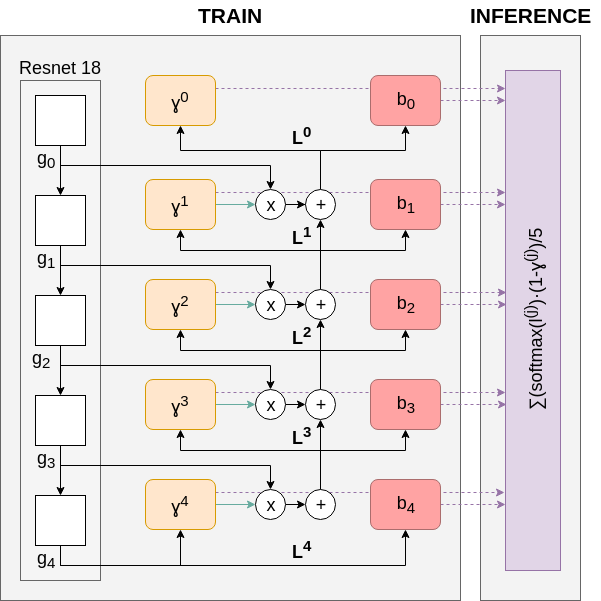}
    \end{center}
  \caption{Uncertainty gated convolutional neural network. Black arrows represent weighted connections between different layers. Green arrows represent forward-only weighted connections, where gradient flows in the backpropagation process are not allowed.}
    \label{fig:ugn}
\end{figure}

Our architecture is illustrated in Figure \ref{fig:ugn}. As it can be seen, an uncertainty measure $\gamma^{(j)}$ is computed after each of the ResNet blocks $g_j$, $0 \leq j \leq 4$. The blocks $g_4$ through $g_1$ correspond to each of the original residual blocks, while $g_0$ is composed by the first max-pooling and convolution. 

The refinement process through uncertainty gates starts by setting $\bar{b}_4 = g_4$ and $b_j$ the upsampled version of $\bar{b}_j$ to match $g_{j-1}$ dimensions. Then for each $j=4,..,1$ the process of obtaining an uncertainty and segmentation is defined as follows:

\begin{fleqn}

\begin{alignat*}{3}
    &l^{(j)} &&= b^{(j)} \circledast_{\mbox{\tiny C}} \mathbf{w^{(j,1)}_{1x1}} && \\
    &\sigma^{(j)} &&= b^{(j)} \circledast_{\mbox{\tiny C}} \mathbf{w^{(j,2)}_{1x1}} && \\
    &\hat{l}^{(j)}_{i,t} &&\sim \mathcal N(l^{(j)}_i, \sigma^{(j)}_i) && \\
    &\gamma^{(j)} &&= \log \frac{1}{T} \sum_t \exp(\hat{l}^{(j)}_{i,t,c} - \log \sum_{c'} \exp \hat{l}^{(j)}_{i,t,c'}) && \\
    &\bar{b}_{j-1} &&= \gamma^{(j)} * g_{j-1} + b_{j}&&
\end{alignat*}
\end{fleqn}

Where $\circledast$\textsubscript{\tiny C} is the convolution operator with a $1\times1$ kernel and dimension $C$, and $*$ denotes the element-wise multiplication, but defined in such a way that gradient can only flow through the $\bar{g_j}$ operand during the backpropagation step. If gradient is allowed to flow through $\gamma^{(j)}$ in the backward pass, we can no longer talk about heteroscedastic uncertainty, as external factors aside from pure classification would condition them. Finally, $\gamma^{(0)}$ and $L^{(0)}$ is computed in the same manner.


\begin{figure*}[h!]
    \captionsetup[subfigure]{labelformat=empty}
    \centering
    \begin{subfigure}[t]{0.13\textwidth}
        \raisebox{-\height}{\includegraphics[width=\textwidth]{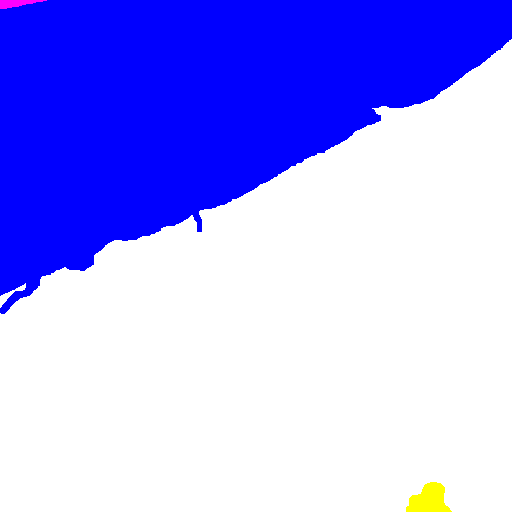}}
        \caption{Ground truth}
    \end{subfigure}
    \hfill
    \begin{subfigure}[t]{0.13\textwidth}
        \raisebox{-\height}{\includegraphics[width=\textwidth]{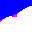}}
        \caption{$b_4$}
    \end{subfigure}
    \hfill
    \begin{subfigure}[t]{0.13\textwidth}
        \raisebox{-\height}{\includegraphics[width=\textwidth]{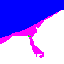}}
        \caption{$b_3$}
    \end{subfigure}
    \hfill
    \begin{subfigure}[t]{0.13\textwidth}
        \raisebox{-\height}{\includegraphics[width=\textwidth]{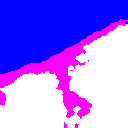}}
        \caption{$b_2$}
    \end{subfigure}
    \hfill
    \begin{subfigure}[t]{0.13\textwidth}
        \raisebox{-\height}{\includegraphics[width=\textwidth]{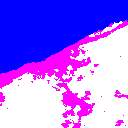}}
        \caption{$b_1$}
    \end{subfigure}
    \hfill
    \begin{subfigure}[t]{0.13\textwidth}
        \raisebox{-\height}{\includegraphics[width=\textwidth]{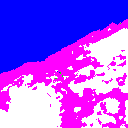}}
        \caption{$b_0$}
    \end{subfigure}
    \hfill
    \begin{subfigure}[t]{0.13\textwidth}
        \raisebox{-\height}{\includegraphics[width=\textwidth]{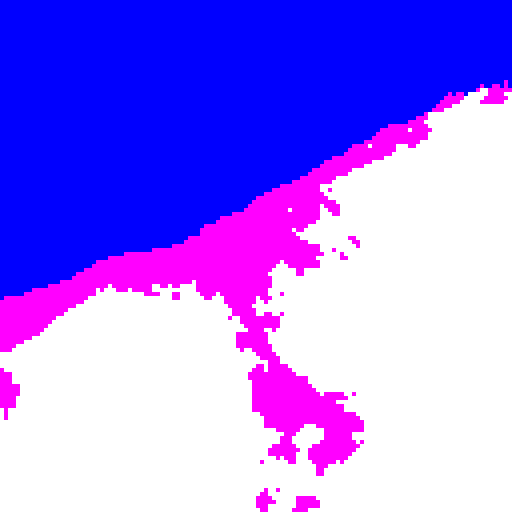}}
        \caption{Output segmentation}
    \end{subfigure}
    \begin{subfigure}[t]{0.13\textwidth}
        \raisebox{-\height}{\includegraphics[width=\textwidth]{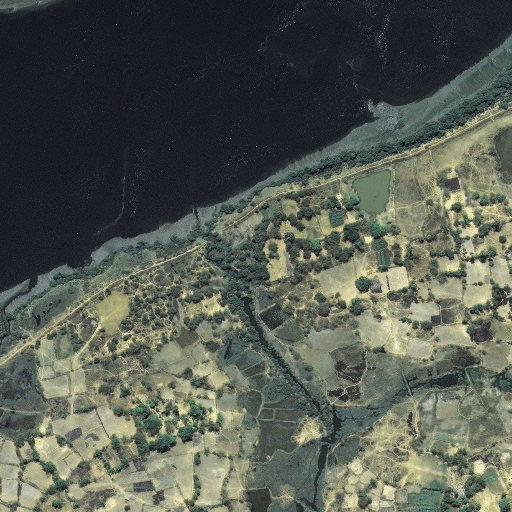}}
        \caption{Input image}
    \end{subfigure}
    \hfill
    \begin{subfigure}[t]{0.13\textwidth}
        \raisebox{-\height}{\includegraphics[width=\textwidth]{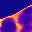}}
        \caption{$\gamma^{(4)}$}
    \end{subfigure}
    \hfill
    \begin{subfigure}[t]{0.13\textwidth}
        \raisebox{-\height}{\includegraphics[width=\textwidth]{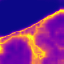}}
        \caption{$\gamma^{(3)}$}
    \end{subfigure}
    \hfill
    \begin{subfigure}[t]{0.13\textwidth}
        \raisebox{-\height}{\includegraphics[width=\textwidth]{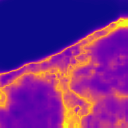}}
        \caption{$\gamma^{(2)}$}
    \end{subfigure}
    \hfill
    \begin{subfigure}[t]{0.13\textwidth}
        \raisebox{-\height}{\includegraphics[width=\textwidth]{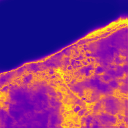}}
        \caption{$\gamma^{(1)}$}
    \end{subfigure}
    \hfill
    \begin{subfigure}[t]{0.13\textwidth}
        \raisebox{-\height}{\includegraphics[width=\textwidth]{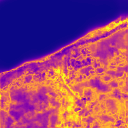}}
        \caption{$\gamma^{(0)}$}
    \end{subfigure}
    \hfill
    \hspace{0.13\textwidth}
    \caption{Prediction process of an image. All $b_j$ are intermediate segmentation outputs, resulting from a $softmax(l^{(j})$, and upsampled to target resolution. Low uncertainty is represented by blue, yellow indicates high uncertainty.}
    \label{fig:pipeline}
\end{figure*}

To compute the final segmentation, instead of taking new logits from the last block as GCN does, it is proposed a method that takes advantage of all logits $l^{(j)}$ and uncertainties $\gamma^{(j)}$ calculated at each block. The final probabilities of each pixel and class are obtained with a $\gamma$-weighted sum of the probabilities at each intermediate step, as:

$$\frac{1}{C}\sum_{i=0}^{4} softmax(l^{(i)}) * (1 - \gamma^{(i)})$$

\section{Training}

To train the model, we first reduce the original resolution down to $1024\times1024$, which simplifies the problem space while keeping enough details. The network is further trained by taking 8 random crops, each of $250\times250$, out of each image. Each crop is then randomly rotated and flipped, and is further processed by adding gaussian noise and adjusting hue, contrast and brightness.

The model is trained by minimizing, at each level $j$, both $L^{(j)}$ and a classification loss given by a softmax crossentropy between the labels and sampled unaries from the logits. Overall, loss is minimized with WNAdam optimizer \cite{wu2018wngrad}, using an standard piecewice learning rate decay for a total of 100 epochs.

\section{Results}

The data for the DeepGlobe Land Cover Classification Challenge consists of 1.146  satellite RGB images of size 2448x2448 pixels, split into training/validation/test, each with 803/171/172 images. Each satellite image is paired with a class labeled image using the following 7 categories: 1) Urban land; 2) Agriculture land; 3) Rangeland; 4) Water (Rivers, oceans, lakes, wetland, ponds); 5) Barren land (Mountain, land, rock, dessert, beach, no vegetation) and 7) Unknown (clouds and other artifacts).

The pixel-wise mean Intersection over Union (mIoU) score, calculated by averaging the IoU over all classes, is used as evaluation metric. The IoU is defined as: True Positive / (True Positive + False Positive + False Negative). The \emph{unknown} class is not an active class used in evaluation.

The final model uses, as discussed in the previous section, the ResNet with 18 layers and is trained for 100 epochs. Figure \ref{fig:pipeline} shows the prediction process. Segmentations at each level are generated for visualization and interpretation and further combined to obtain the final result. Deeper levels are more general and can not accurately predict each pixel, which can be both attributed to the downsampling process and the abstraction done through all the convolutions. That is why upper levels refine the result and are richer in details. In particular, it can be seen that pixels where the output does not match the ground truth, a high uncertainty is obtained. Averaging across all levels improves the result by reducing artifacts and producing smoother segmentation maps. The model runs inference in real time, taking only 250ms to produce a segmentation at full $2448\times2448$ resolution on an NVIDIA Titan X. This architecture achieves a mIoU score of 0.485 in the final test set of the challenge.

\section{Conclusions}
In this paper, an uncertainty gated convolutional neural network has been proposed for land-cover semantic segmentation. The proposed method leverages the computation of a heteroscedastic measure of uncertainty when classifying individual pixels in an image. This classification uncertainty measure is used to define a set of memory gates between layers that allow for a principled method to select the optimal decision for each pixel. The result reported on the DeepGlobe Land Cover Classification Challenge is 0.485 mIoU on the final test set. Future improvement on the domain adaptation problem will be considered, since we have observed some inconsistencies due to this specific issue.

\ifcvprfinal\
\section*{Acknowledgements}
This work was partially founded by MINECO Grant TIN2015-66951-C2 and by an FPU grant (Formación de Profesorado Universitario) from the Spanish Ministry of Education, Culture and Sport (MECD) to Guillem Pascual (FPU16/06843). We gratefully acknowledge the support of NVIDIA Corporation with the donation of the Titan X Pascal GPU used for this research. 
\fi

{\small
\bibliographystyle{ieee}
\bibliography{egbib}
}

\end{document}